\documentclass[letterpaper, 10 pt, conference]{ieeeconf}  

\IEEEoverridecommandlockouts                              

\usepackage{graphicx} 
\usepackage{subcaption} 
\usepackage{cite}
\usepackage{caption}
\usepackage{booktabs}
\usepackage{booktabs} 
\usepackage{tabularx}
\usepackage{makecell}
\usepackage{longtable}
\usepackage{multirow} 
\usepackage{array} 
\usepackage{dirtree}
\usepackage{placeins}
\usepackage{amsmath} 
\usepackage{amsmath} 
\usepackage{amssymb}  
\usepackage{fancyhdr}

\usepackage{algorithm}
\usepackage{algpseudocode}
\usepackage{amsmath}

\title{\LARGE \bf
A High-accuracy Event-based Underwater SLAM System
}

\author{Yifan Peng*, Qihang Liu*, Haoying Li, Yuzhe Li, Junfeng Wu, Ziyang Hong}

\begin{document}

\maketitle
\thispagestyle{empty}
\pagestyle{empty}

\begin{abstract}

While event cameras offer immense potential for underwater SLAM, existing Time Surface (TS)-based methods prove highly unreliable when deployed underwater. Fluctuating camera velocities severely degrade TS imaging quality, while wide stereo baselines and repetitive underwater textures induce critical matching failures, frequently triggering system failure. To overcome these challenges, we develop the first high-accuracy event-based underwater stereo SLAM system. A structure-aware metric for TS is designed based on structure tensor coherence and gradients to quantitatively evaluate TS structural information density. By decoupling the optimal TS generation into two distinct stages based on system initialization, Bayesian Optimization(BO) first predicts an optimal prior TS sequentially before initialization while we set an asynchronous online local searching method periodically to obtain appropriate TS in real-time during the tracking stage. We use the prior disparity to guarantee precise data association and ``latest-observation-first'' triangulation mechanism to realize stable triangulation. As a benchmark for these solutions and a resource for the community, we also contribute UWE, the first high-quality real-world underwater event dataset containing variable camera motions, complex textures and different trajectory features. Extensive evaluations on public datasets and UWE show the competitive accuracy performance of the proposed SLAM system compared to the state-of-the-art event-based method. The code and data will be open-sourced.

\end{abstract}


\section{Introduction}


As underwater robotics evolves from remotely operated vehicles toward autonomy, reliable perception remains a critical bottleneck. In GPS-denied underwater environments, dynamic illumination and complex textures frequently cause traditional visual systems to fail. By leveraging the microsecond-level latency and high dynamic range (HDR) of event cameras, event-based SLAM holds potential for underwater perception.

Although learning-based event processing shows promise \cite{klenk2024deep, guan2025deio}, synchronous 2D representations, specifically the Time Surface (TS), remain prevalent in onboard SLAM. By encoding the temporal recency of events into an image-like format, TS elegantly bridges asynchronous data with mature visual front-ends under strict real-time constraints \cite{chen2023esvio, niu2025esvo2}. However, deploying TS-based systems underwater reveals notable limitations. First, TS imaging quality is highly sensitive to the varying textures and fluctuating velocities typical of aquatic environments, making it susceptible to feature sparsity or motion blur. Second, the wide stereo baseline required for accurate depth estimation, combined with abundant repetitive underwater textures, frequently induce severe spatial mismatches. Such unreliable data association severely compromises overall system stability. Furthermore, the field lacks comprehensive real-world underwater event datasets with precise ground truth, as current benchmarks primarily target terrestrial domains.

To address these issues, this paper presents systematic innovations spanning from front-end event representation to the geometric backend. Inspired by traditional automatic exposure control \cite{kim2018exposure}, we treat the time-decay coefficient as the ``exposure time'' and design the first information metric for event-based TS to quantitatively evaluate TS image quality and maintain optimal imaging quality across varying motions and textures. Furthermore,an adaptive disparity prior is leveraged to enhance stereo tracking accuracy. Additionally, a ``latest-observation-first'' depth propagation strategy is proposed to resolve data association degradation and initialization instability in complex underwater environments, achieved by maximizing the utilization of historical feature trajectories.
These algorithms are tightly integrated into an accurate event SLAM system, accompanied by the release of an underwater dataset with high-precision groundtruth.

\begin{figure*}[htbp]
    \centering
    \includegraphics[width=\linewidth]{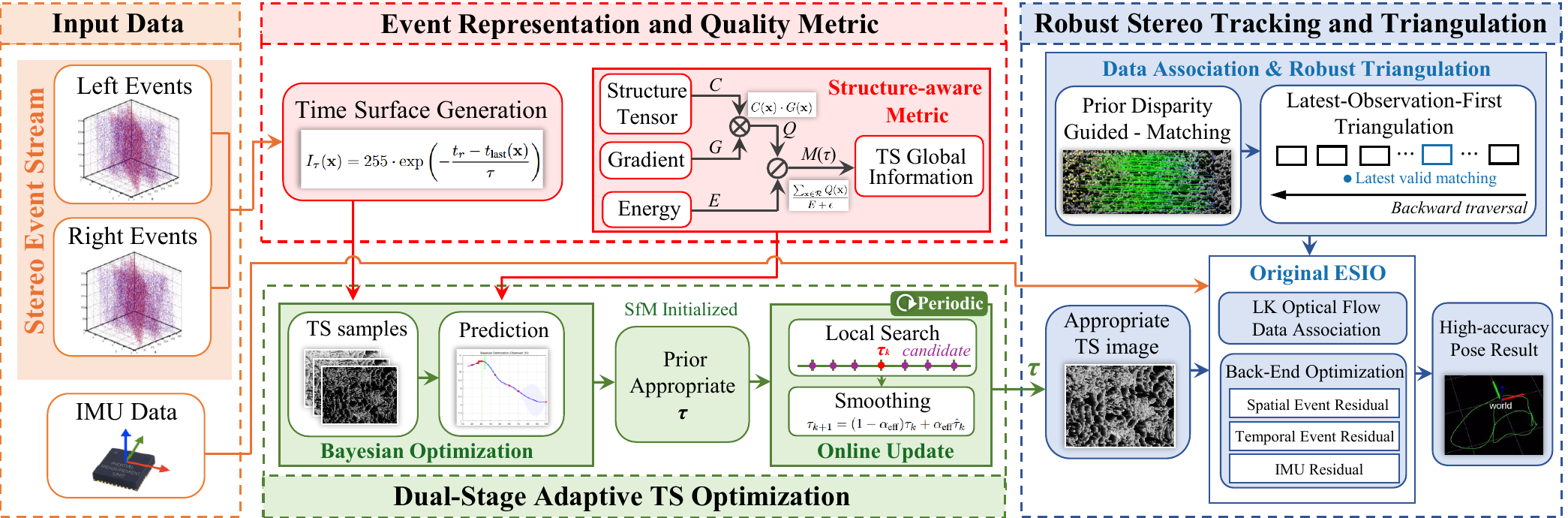}
    \captionsetup{justification=justified,singlelinecheck=false}
    \caption{Overview of the proposed underwater stereo event-based SLAM system. The pipeline developed on ESIO consists of three core modules: (1) \textbf{Event Representation and Quality Metric}, which evaluates 2D Time Surface images using a novel structure-aware metric; (2) \textbf{Dual-Stage Adaptive TS Optimization}, employing BO to predict an optimal time-decay constant ($\tau$) for SfM initialization, followed by an asynchronous local search for online $\tau$ updating; and (3) \textbf{Robust Stereo Tracking and Triangulation}, which integrates a median-based disparity prior to guide LK optical flow and a ``latest-observation-first'' mechanism to robustly recover features from initial matching failures.}
    \label{f1}
\end{figure*}

The paper's main contributions are as follows:

\begin{itemize}

    \item A dynamic framework is proposed to explicitly evaluate TS imaging quality and adjust the time-decay coefficient in real time. Specifically, a \textbf{structure-aware information metric} is designed based on structure tensors and gradient magnitudes. Accompanied by a \textbf{dual-stage adaptive optimization algorithm} synergizing Bayesian Optimization (BO) \cite{snoek2012practical} and online local search, this mechanism resolves representation degradation under varying dynamics without compromising real-time tracking efficiency.

    \item A robust stereo tracking and dynamic triangulation strategy is developed to tackle large disparities and repetitive textures in underwater environments. It introduces a historical median-based \textbf{adaptive disparity prior} to enhance optical flow matching accuracy, and a \textbf{latest-observation-first} depth propagation mechanism that breaks conventional first-frame dependency, substantially boosting initialization and backend stability.

    \item The first underwater pure-event stereo SLAM system integrating the above contributions is developed and open-sourced. Furthermore, a challenging high-quality \textbf{U}nder\textbf{W}ater \textbf{E}vent Dataset(UWE) featuring diverse motions, complex textures, and extreme lighting is released. Extensive evaluations on public benchmarks and UWE demonstrate the system's superior accuracy, robustness, and real-time performance.
    
\end{itemize}

\section{Related Work}

\subsection{Event-based SLAM}
Terrestrial event SLAM has been extensively studied, with the optimization of 2D event representations emerging as a critical research direction. Unlike early qualitative TS evaluations \cite{mueggler2015lifetime, manderscheid2019speed}, recent works integrate dynamic adaptations directly into SLAM pipelines. For instance, \cite{jiao2021comparing} dynamically switches to an Event Map during event sparsity using Hessian eigenvalues, while \cite{liu2023t} adaptively tunes the time-decay parameter based on event counts to mitigate degradation under varying speeds. Regarding complete open-source systems, \cite{chen2023esvio} employs IMU-guided motion compensation to sharpen edges for tightly-coupled corner tracking, and \cite{niu2025esvo2} introduces an Adaptive Accumulation (AA) map for efficient contour extraction. However, these terrestrial frameworks overlook the repetitive textures and initialization instability inherent to underwater environments, and the excessive parameterization of \cite{niu2025esvo2} severely hinders its underwater adaptation.

Meanwhile, underwater event SLAM remains largely unexplored. Although \cite{fan2024underwater} pioneered this field by incorporating motion-compensated event frames, the system fundamentally relies on standard cameras and Doppler Velocity Logs (DVLs), treating event reprojection errors merely as an auxiliary loss. Furthermore, its experiments are restricted to simplified settings and lack precise motion-capture ground truth for rigorous quantitative validation.

\subsection{Underwater Datasets}
Currently, the majority of underwater datasets are used for computer vision tasks, such as semantic information extraction and optical flow tracking \cite{ziqiang2024hkcoral, islam2020semantic, truong2026ueof}, whereas datasets dedicated to underwater SLAM remain exceedingly scarce. For instance, although the dataset in \cite{rahman2018sonar} can be utilized to validate Visual-Inertial Odometry (VIO) systems, it lacks ground-truth trajectories. FLsea \cite{randall2023flsea} is a comprehensive, large-scale, real-world underwater dataset providing images and IMU measurements, but it is primarily designed for depth estimation tasks. While it can be repurposed for SLAM evaluation, its authors explicitly acknowledge that the provided ground truth ``may contain slight imperfections.'' Although AquaticVision \cite{peng2025aquaticvision} is the pioneering underwater dataset providing synchronized events, frames, and motion-capture ground truth, its excessively slow camera motion results in sparse event volumes. This significantly compromises data utility, rendering it insufficient for rigorously evaluating the robustness of event-based SLAM systems.

\section{Methodology}

\subsection{Overview} \label{ov}
We build our underwater stereo event-based SLAM system upon the ESIO framework \cite{chen2023esvio}. As shown in Fig. \ref{f1}, raw event streams are first converted into 2D TS images, whose quality is evaluated by the designed structure-aware metric. The TS evaluation comprises two stages, demarcated by the Structure-from-Motion (SfM) initialization. Prior to initialization, BO leverages the proposed metric to predict an appropriate time decay constant ($\tau$) prior, which remains fixed during SfM for numerical stability. Once initialized, the system enters the continuous SLAM phase, where $\tau$ optimization is decoupled into an asynchronous thread. This thread periodically updates $\tau$ online by local searching with the prior $\tau$ to generate high-quality TS images for robust front-end data association. To further enhance tracking accuracy, a median-based disparity prior is introduced to guide Lucas-Kanade (LK) optical flow \cite{lucas1981iterative}. Concurrently, a ``latest-observation-first'' mechanism mitigates initial matching failures by reversely traversing historical feature trajectories. This identifies the most recent valid stereo match for high-confidence triangulation. Ultimately, these integrated strategies guarantee precise initialization and significantly boost the overall pose estimation accuracy and robustness.

\subsection{Event Representation and TS Quality Metric}

\subsubsection{TS Image Generation}
Event cameras output an asynchronous stream \(\mathcal{E} = \{e_k\}_{k=1}^N\), where \(e_k = (\mathbf{x}_k, t_k, p_k)\) denotes pixel coordinates, timestamp, and polarity. The TS image, as a commonly used representation method in event SLAM, can reflect the temporal recency of event triggers. At any given reference time $t_r$, historical events are projected onto a 2D plane following an exponential decay rule \cite{7508476}. For a pixel location $\mathbf{x} = (u, v)$, its intensity $I_\tau(\mathbf{x})$ is defined as:
\begin{equation}
    I_\tau(\mathbf{x}) = 255 \cdot \exp\left(-\frac{t_r - t_{\text{last}}(\mathbf{x})}{\tau}\right), \quad t_r \ge t_{\text{last}}(\mathbf{x}),
    \label{eq:ts_generation}
\end{equation}
where $t_{\text{last}}(\mathbf{x})$ is the timestamp of the most recent event triggered at this pixel. The time-decay coefficient $\tau$ directly determines the lifespan of historical events on the current surface.

\subsubsection{Structure-Aware Information Metric}

\begin{figure}[htbp]
    \centering
    \includegraphics[width=\linewidth]{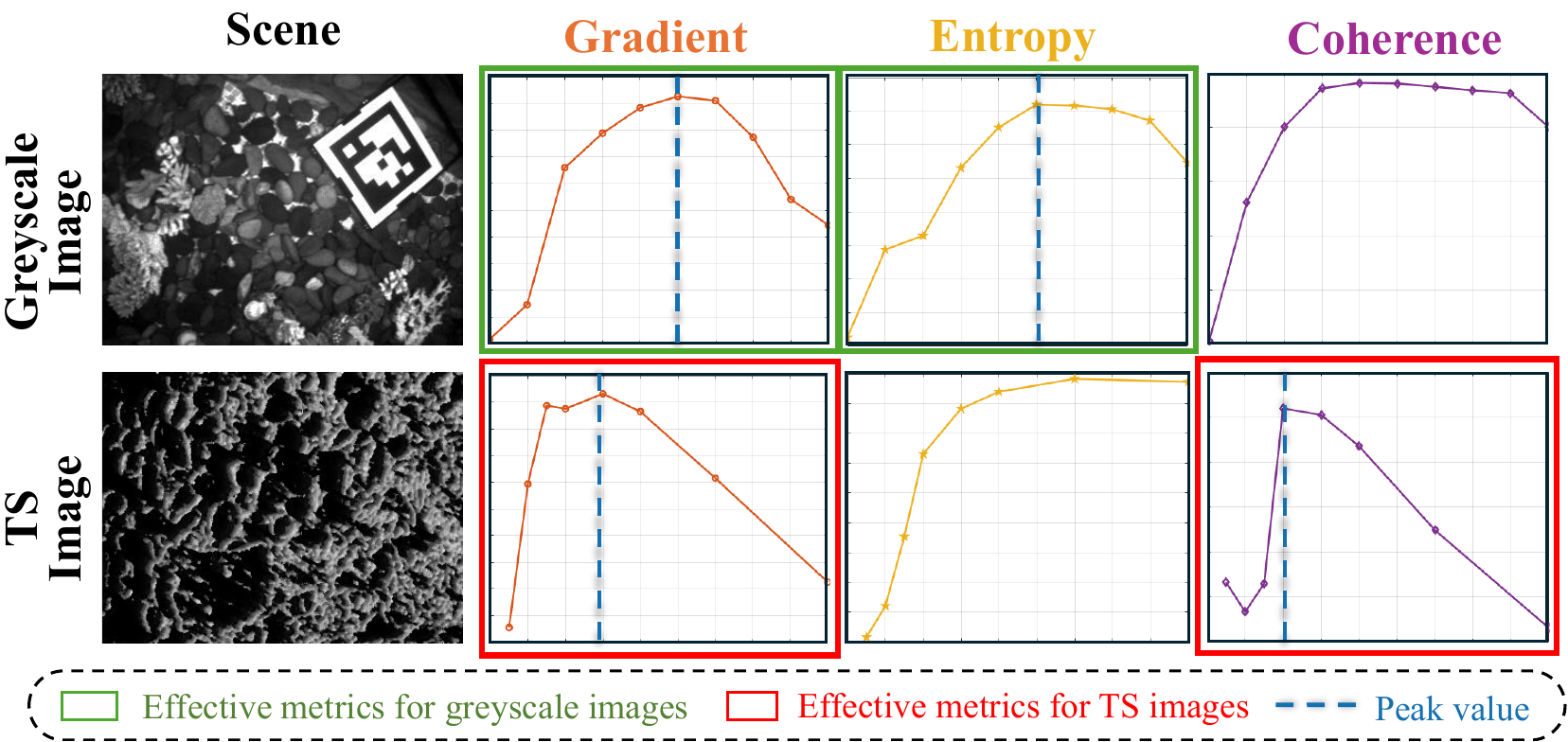}
    \captionsetup{justification=raggedright,singlelinecheck=false}
    \caption{Comparison of statistical feature evolution between traditional optical images and event-based TS images.}
    \label{f2}
\end{figure} 

According to \eqref{eq:ts_generation}, overall TS intensity increases monotonically with $\tau$, functioning analogously to camera exposure time. Inspired by auto-exposure frameworks \cite{kim2018exposure}, we design a quantitative metric to dynamically identify the optimal TS. Crucially, a valid objective function must exhibit a distinct global maximum. As Fig. \ref{f2} illustrates, while traditional image statistics (entropy) peaks at optimal exposure, TS behavior fundamentally differs. As historical events accumulate, TS entropy increases monotonically with $\tau$, rendering it mathematically unsuitable for optimization. Conversely, spatial gradient and structure tensor coherence demonstrate well-defined peaks, perfectly encapsulating the physical trade-off between sufficient event energy and minimal motion blur. Therefore, synthesizing gradient and coherence forms the theoretical foundation of our evaluation metric.

The structure tensor $\mathbf{J}_\rho(\mathbf{x})$ captures the predominant gradient directions within a neighborhood \cite{weickert2002scheme}. It is computed by convolving the outer product of the normalized tensor gradient $\nabla I_\tau \nabla I_\tau^T$ with a Gaussian kernel $G_\rho$ of integration scale $\rho$, where $\rho > \sigma_g$ and $\sigma_g$ is the scale of the gradient variations:

\begin{equation}
    \mathbf{J}_\rho(\mathbf{x}) = G_\rho \circledast \left(\nabla I_\tau \nabla I_\tau^T\right) = 
    \begin{bmatrix} 
        J_{uu} & J_{uv} \\ 
        J_{uv} & J_{vv} 
    \end{bmatrix}.
\end{equation}

The tensor’s eigenvalues $\lambda_1$ and $\lambda_2$ ($\lambda_1 \geq \lambda_2 \geq 0$) characterize the local directional structure, where $\lambda_1$ represents the dominant gradient direction. A large $\lambda_1 / \lambda_2$ ratio indicates a strong, unidirectional edge, as in sharp edges where $\lambda_1 \gg \lambda_2$.

The coherence metric $C(\mathbf{x})$ quantifies edge sharpness is defined as:

\begin{equation}
    C(\mathbf{x}) = \left( \frac{\lambda_1 - \lambda_2}{\lambda_1 + \lambda_2 } \right)^2.
\end{equation}
Sharp edges correspond to high coherence, while blurred edges lead to low coherence, $C(\mathbf{x}) \approx 0$.

To prevent the events from being overly sparse and the overall TS energy from being too low while meeting the sharpness criteria, the gradient magnitude $G(\mathbf{x})= || \nabla I_\tau ||$ is further incorporated. Therefore, we define the structure quality score for a single valid pixel as:
\begin{equation}
    Q(\mathbf{x}) = C(\mathbf{x}) \cdot G(\mathbf{x}).
\end{equation}

To eliminate the interference of static background noise in event-free regions, a valid pixel set with intensities greater than a small threshold $\eta$ is extracted, denoted as $\mathcal{R} = \{\mathbf{x} \mid \hat{I}_\tau(\mathbf{x}) > \eta\}$. Under the current decay coefficient $\tau$, the global structure-aware score metric $M(\tau)$ for the entire TS is defined as:
\begin{equation}
    M(\tau) =\frac{\sum_{\mathbf{x} \in \mathcal{R}} Q(\mathbf{x})}{\sum_{\mathbf{x} \in \mathcal{R}} I_\tau(\mathbf{x})},
\end{equation}
where denominator represents the total valid energy of the TS. 
$M(\tau)$ represents the structural information density per unit energy. A larger value of $M(\tau)$ indicates a more appropriate TS, because its energy is highly concentrated on meaningful geometric structures rather than being dispersed.

\subsection{Dual-Stage Adaptive Time-Decay Optimization}
To get an appropriate $\tau^*$ that produces the best TS, we formulate the problem as maximizing the structure-aware score metric $M(\tau)$:
\begin{equation}
    \tau^* = \arg\max_{\tau \in [\tau_{\min}, \tau_{\max}]} M(\tau),
\end{equation}
where $\tau_{\min}$ and $\tau_{\max}$ represent the empirically determined lower and upper bounds for the time-decay coefficient.

In experiments, we found that frequently updating $\tau$ degrades the system's accuracy and robustness. This occurs mainly because: (1) During camera turns, the velocity briefly drops to zero, causing the production of inaccurately large $\tau$ values. When motion resumes, these aberrant values act as noise, which is particularly detrimental during the sensitive initialization phase and can lead to system failure. (2) Computing a new $\tau$ for every frame introduces significant latency, impairing real-time performance and further undermining stability. Therefore, as detailed in section~\ref{ov} and the green part of Fig. \ref{f1}, we decompose the adaptive TS optimization into two distinct stages with different methods.

\subsubsection{Sequential Bayesian Optimization}

During the system startup stage, BO and SfM execute sequentially. BO first predicts an optimal $\tau$ while the system remains in a waiting state. Only after this prediction is obtained does SfM initialization commence. The value of $\tau$ is kept fixed until the initialization is successfully completed.

Since finding the optimal $\tau$ in the global space is a black-box optimization problem, Gaussian Process (GP) \cite{seeger2004gaussian} is utilized as the surrogate model for the objective function $M(\tau)$:
\begin{equation}
    f(\tau) \sim \mathcal{GP}(m(\tau), k(\tau, \tau')),
\end{equation}
where the mean function $m(\tau)$ is set to $0$. The covariance function $k(\tau, \tau')$ employs a Radial Basis Function (RBF) to measure the correlation between different $\tau$ values. 

Assuming $n$ sampling evaluations have been performed, an observation set is obtained:
\begin{equation}
    \mathcal{D}_n = \{(\tau_i, y_i)\}_{i=1}^n,
\end{equation}
where:
\begin{equation}
    y_i = M(\tau_i) = f(\tau_i) + \epsilon_i, \quad \epsilon_i \sim \mathcal{N}(0, \sigma_\epsilon^2),
\end{equation}
with $\epsilon_i$ representing the observation noise. For any unknown test point $\tau_*$ within the global space, its posterior predictive distribution follows a normal distribution:
\begin{equation}
    \mathcal{N}(\mu_i(\tau_*), \sigma_i^2(\tau_*)).
\end{equation}

The analytical solutions for its mean and variance are given by:
\begin{equation}
    \mu_n(\tau_*) = \mathbf{k}_*^T (\mathbf{K} + \sigma_\epsilon^2 \mathbf{I})^{-1} \mathbf{y},
\end{equation}
\begin{equation}
    \sigma_n^2(\tau_*) = k(\tau_*, \tau_*) - \mathbf{k}_*^T (\mathbf{K} + \sigma_\epsilon^2 \mathbf{I})^{-1} \mathbf{k}_*,
\end{equation}
where $\mathbf{K} \in \mathbb{R}^{n \times n}$ is the covariance matrix among existing observation points, $\mathbf{y} \in \mathbb{R}^n$ is the column vector of observed values, $\mathbf{k}_* \in \mathbb{R}^n$ is the covariance vector between the test point $\tau_*$ and all existing observations, and $\mathbf{I}$ is the identity matrix.

To determine the optimal $\tau_{n+1}$ for the next evaluation, we find the optimal $\tau$ that maximizes the Expected Improvement (EI) \cite{jones1998efficient} acquisition function:
\begin{equation}
    \alpha_{\text{EI}}(\tau) = (\mu_n(\tau) - f^+) \Phi(Z) + \sigma_n(\tau)\phi(Z),
\end{equation}
\begin{equation}
    Z = \frac{\mu_n(\tau) - f^+}{\sigma_n(\tau)},
\end{equation}
where $f^+$ is the maximum observed value, and $Z$ represents the normalized improvement that balances exploration and exploitation. $\Phi(\cdot)$ and $\phi(\cdot)$ are the standard normal CDF and PDF respectively. This iterative optimization ensures rapid convergence to the optimal time-decay coefficient $\tau^*$ with minimal evaluations. 

Sequential BO provides a tuning-free, high-quality event representation for SfM initialization, with its output directly serving as the prior for the subsequent local search.

\subsubsection{Asynchronous Online Local Searching and Smoothing}

\begin{figure}[htbp]
    \centering
    \includegraphics[width=\linewidth]{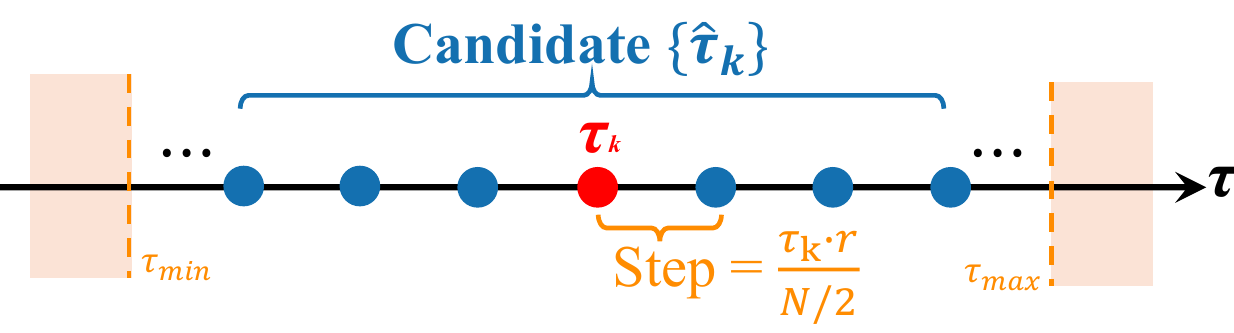}
    \captionsetup{justification=raggedright,singlelinecheck=false}
    \caption{Discrete local search for online $\tau$ optimization. N is the size of candidate set.}
    \label{f3}
\end{figure}

Following SfM initialization, $\tau$ optimization transitions to an independent thread to prevent tracking latency. As illustrated in Fig. \ref{f3}, an adaptive discrete local search is periodically executed. A symmetric candidate set is dynamically sampled around the currently active $\tau$ using a predefined search ratio $r$. The candidate hitting the maximum score $M(\tau)$ is then selected as the local optimum. 

To prevent the failure of Lucas-Kanade (LK) optical flow tracking caused by the violation of photometric consistency arising from drastic variations in $\tau_k$,the current $\tau_k$ is smoothly updated to its next state $\tau_{k+1}$ using the local optimum $\hat{\tau}_k$ via an adaptive first-order low-pass filter:
\begin{equation}
    \tau_{k+1} = (1 - \alpha_{\text{eff}}) \tau_k + \alpha_{\text{eff}} \hat{\tau}_k,
\end{equation}
where $\alpha_{\text{eff}}$ is the adaptive decay factor designed to mitigate noise-induced parameter oscillations:
\begin{equation}
    \alpha_{\text{eff}} = \begin{cases} 
        \alpha, & M(\hat{\tau}_k) \ge M(\tau_k) - \delta \\ 
        \gamma \alpha, & M(\hat{\tau}_k) < M(\tau_k) - \delta ,
    \end{cases}
\end{equation}
where $\alpha \in (0, 1)$ is the base smoothing coefficient and $\delta$ is the significance threshold. The physical rationale is straightforward: the update proceeds with the normal step size $\alpha$ only if the new candidate $\hat{\tau}_k$ offers a notable quality improvement or remains stable. Otherwise, the update step is deliberately scaled down by the suppression coefficient $\gamma \in (0, 1)$ to maintain current stability.

In summary, sequential BO ensures a robust cold start, while asynchronous local searching maintains continuous adaptability without compromising real-time tracking. This complementary design enhances 2D event representation in event-based SLAM across varying speeds and scene textures.


\subsection{Robust Stereo Tracking and Triangulation}

Unstructured underwater environments with repetitive textures, combined with the wide baselines required for accurate stereo depth estimation, induce extreme pixel disparities within the TS representation. Optimizing only the time-decay coefficient $\tau$ improves single-camera temporal tracking but fails to resolve stereo LK matching failures. 

\subsubsection{Adaptive Disparity-Guided Stereo Matching}
To address LK optical flow limitations under large disparities and repetitive textures, we propose an adaptive disparity prior-guided algorithm. Utilizing a historical median-based disparity range provides a reliable initial translation for the stereo search, significantly boosting feature matching success.

Let $\{\mathbf{p}_{l,i}^k\}_{i=1}^M$ denote the set of successfully extracted and tracked feature points on the left TS at time step $k$, with pixel coordinates expressed as $\mathbf{p}_{l,i}^k = [u_{l,i}^k, v_{l,i}^k]^T$. Their corresponding matched points in the right camera are denoted as $\{\mathbf{p}_{r,i}^k\}_{i=1}^M = \{[u_{r,i}^k, v_{r,i}^k]^T\}_{i=1}^M$. For any successfully matched feature $i$, its 2D spatial disparity vector is defined as:
\begin{equation}
    \mathbf{d}_i^k = \mathbf{p}_{l,i}^k - \mathbf{p}_{r,i}^k = \begin{bmatrix} d_{u,i}^k \\ d_{v,i}^k \end{bmatrix}.
\end{equation}

During system operation, the disparities of all successfully matched feature points from the previous frame are continuously recorded. To prevent anomalous disparities caused by occasional false matches, the median of these historical disparities is extracted to serve as the global default prior $\bar{\mathbf{d}}^k$ for the current frame.

For each target feature point $i$ to be matched in the current frame, an individual prior $\hat{\mathbf{d}}_i^k$ is assigned using a ``history-first'' strategy. If the ID of the feature exists in the previous frame's mapping set $\mathcal{D}_{k-1}$, its own historical disparity is reused; otherwise, the global median prior is utilized:
\begin{equation}
    \hat{\mathbf{d}}_i^k = \begin{cases} 
        \mathcal{D}_{k-1}(\text{id}_i), & \text{if } \text{id}_i \in \mathcal{D}_{k-1} \\ 
        \bar{\mathbf{d}}^k, & \text{otherwise} 
    \end{cases}.
\end{equation}

Based on this, the proposed algorithm performs prior-guided matching and consistency filtering on the feature points. Before executing the LK optical flow, a spatial offset is applied to the search starting point in the right view using the prior $\hat{\mathbf{d}}_i^k$:
\begin{equation}
    \tilde{\mathbf{p}}_{r,i}^k = \mathbf{p}_{l,i}^k - \hat{\mathbf{d}}_i^k.
\end{equation}

Subsequently, using $\tilde{\mathbf{p}}_{r,i}^k$ as the initial guess, sub-pixel optical flow optimization is executed on the right TS to obtain the final matched point $\mathbf{p}_{r,i}^k$. To further eliminate false matches caused by highly repetitive textures, a forward-backward consistency check is enforced on the matching results. Only features satisfying a backward tracking error of $\|\mathbf{p}_{l,i}^k - \hat{\mathbf{p}}_{l,i}^k\|_2 < \epsilon_{fb}$ are retained. Finally, these high-confidence disparity results are written back to the mapping set $\mathcal{D}_k$, forming a robust closed loop of adaptive priors.

\begin{figure}[htpb]
    \centering
    \includegraphics[width=\linewidth]{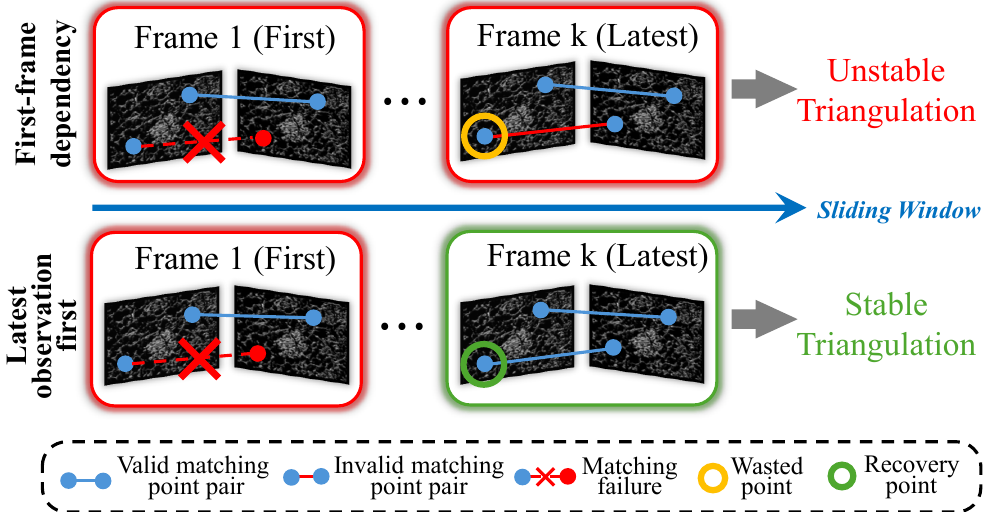}
    \captionsetup{justification=raggedright,singlelinecheck=false}
    \caption{Comparison of feature depth initialization strategies within the sliding window. Our method increases successfully triangulated features compared to ESIO.}
    \label{f4}
\end{figure}

\subsubsection{Latest-Observation-First Triangulation}

The original ESIO framework has a rigid first-frame dependency, namely, a feature's depth initialization relies exclusively on the success of stereo matching at the exact moment it is first extracted. However, in underwater scenarios with large disparities and repetitive textures, these initial observations are highly unstable. Once the first-frame matching fails, the feature's depth remains permanently uninitialized. Even if the feature is stably tracked in subsequent frames, it cannot be triangulated and is eventually discarded, severely depleting the active feature set and triggering SfM collapse.

To address initial matching failures that prematurely discard valid features, a ``latest-observation-first'' mechanism is proposed to recover feature points in subsequent frames for depth initialization. As shown in Fig. \ref{f4}, unlike the rigid first-frame dependency (top), our mechanism (bottom) successfully recovers features that failed initial matching by exploiting their latest valid stereo observations. For a feature trajectory continuously observed across multiple steps in the sliding window, its history is traversed in reverse to identify the most recent valid stereo match (at frame $k^*$). Leveraging this high-quality recent observation, Singular Value Decomposition (SVD)-based linear triangulation is executed:
\begin{equation}
    \mathbf{X}_i = \text{Triangulate}(\mathbf{P}_l^{k^*}, \mathbf{P}_r^{k^*}, \mathbf{p}_l^{k^*}, \mathbf{p}_r^{k^*}).
\end{equation}

Strict geometric rejection criteria are subsequently applied. The depth must fall within $[z_{\min}, z_{\max}]$ and reprojection errors must be sufficiently small for acceptance by the backend. Once validated, these high-confidence features are integrated into the sliding window, propagating their depths during marginalization. By leveraging historical trajectories alongside optimal 2D TS representations, this dynamic initialization ensures robust SfM and stable backend optimization for event-based stereo SLAM in challenging underwater environments.

\section{Experiment} \label{sec:experimental}

\subsection{\textbf{U}nder\textbf{W}ater\textbf{E}vent (UWE) Dataset}

Given the extreme scarcity of datasets dedicated to underwater event SLAM, we open-source the first comprehensive underwater event dataset featuring diverse lighting conditions, variable texture details, and a wide range of motion velocities. This dataset not only serves to validate our proposed system but also aims to foster the development of the underwater event-vision community.

\begin{figure}[htbp]
    \centering
    \includegraphics[width=\linewidth]{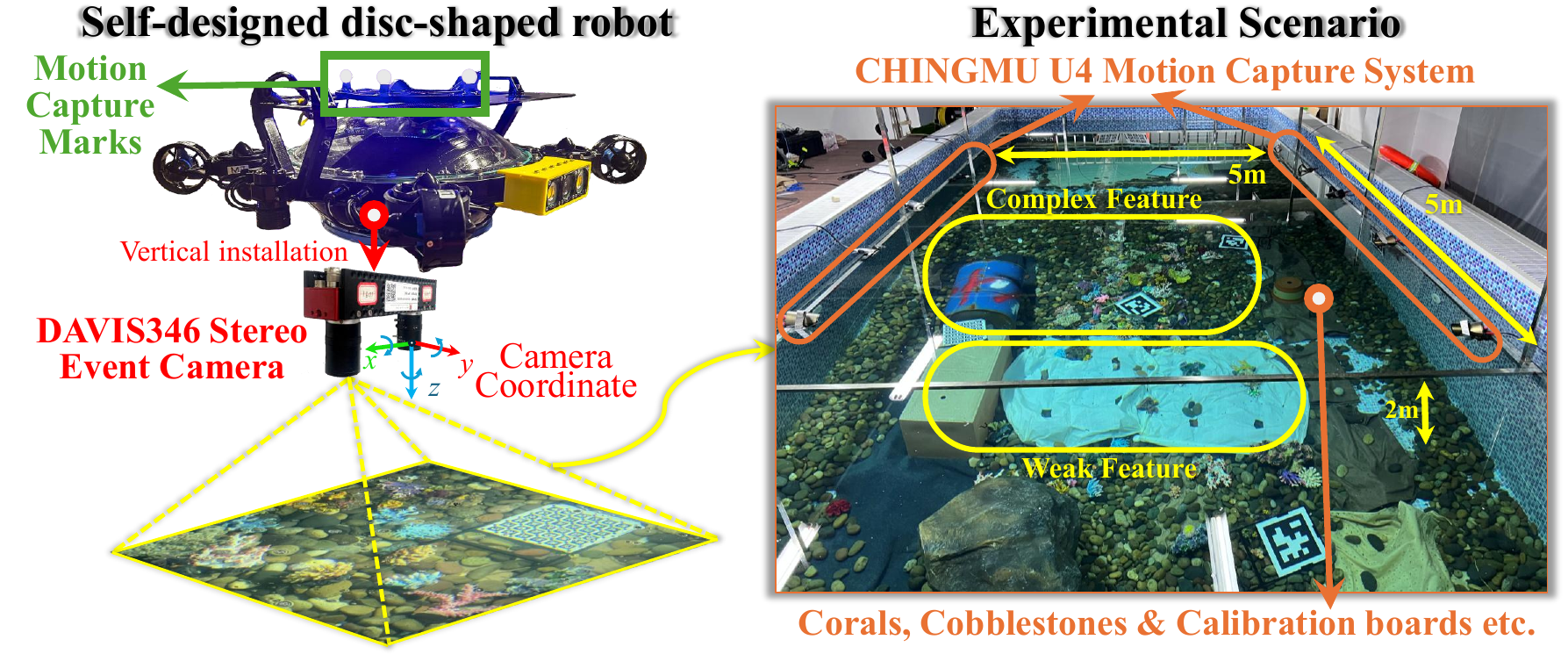}
    \captionsetup{justification=raggedright,singlelinecheck=false}
    \caption{The disc-shaped robot and expreimental scenario.}
    \label{f5}
\end{figure}

\begin{figure*}[htpb]
    \centering
    \includegraphics[width=\linewidth]{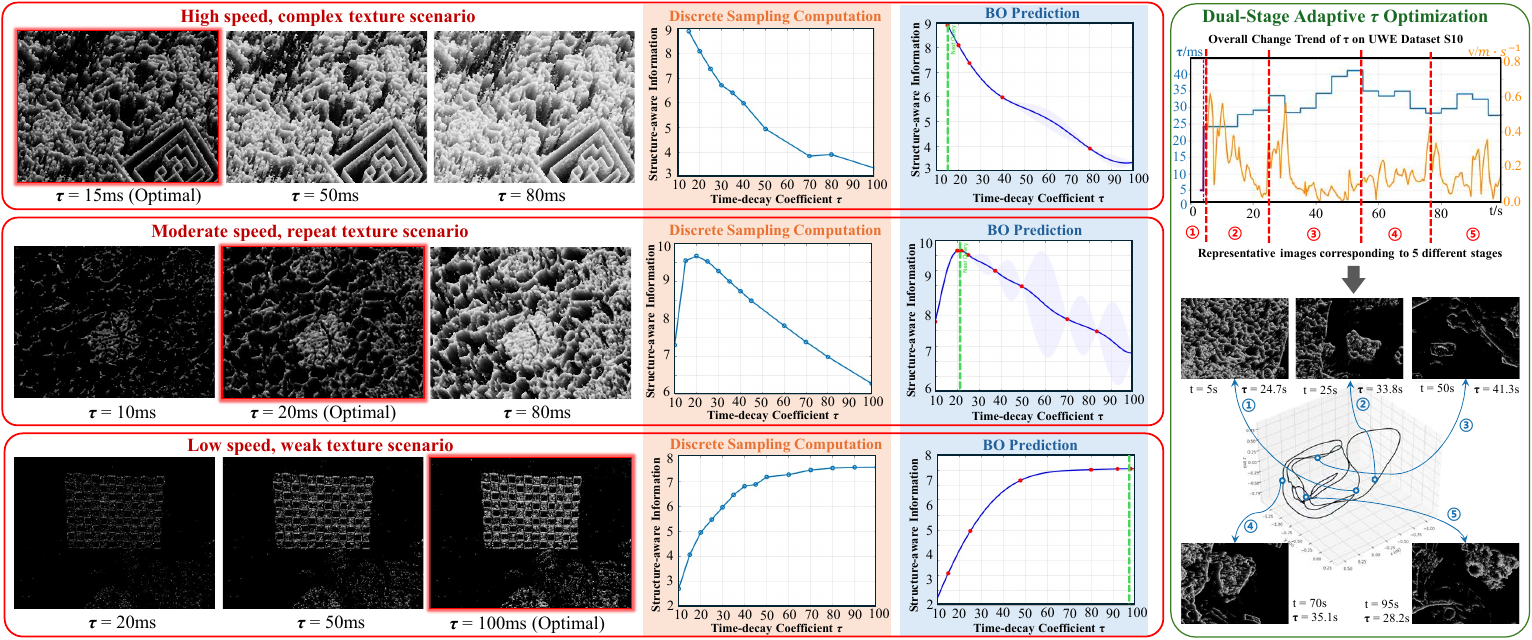}
    \captionsetup{justification=raggedright,singlelinecheck=false}
    \caption{Qualitative and quantitative evaluations of the dual-stage adaptive time-decay optimization algorithm.}
    \label{f6}
\end{figure*} 

Fig. \ref{f5} shows our self-designed underwater robot and a $5 \times 5 \times 2 \text{ m}^3$ pool where the bottom is arranged with corals, cobblestones, calibration boards, and oil drums to construct a texture-diverse underwater environment. Equipped with a downward-facing DAVIS346 stereo event camera, our highly agile disc-shaped robot outperforms traditional ROVs by maneuvering freely across varying speeds, effectively exploiting the high-speed capabilities of event cameras.

The UWE dataset comprises 10 sequences with $60$ Hz events, $1000$ Hz IMU, and $20$ Hz images. The CHINGMU U4 motion capture system provides $120$ Hz groundtruth trajectories. Surpassing the AQV dataset \cite{peng2025aquaticvision}, UWE features greater event volumes, broader velocity profiles, and extended durations ($90$--$150$s). The characteristics of 10 sequences in UWE is concluded in Table \ref{t1}. To our knowledge,  this is the first comprehensive dataset dedicated to underwater event perception containing diverse lighting conditions (well-lit to unlit), challenging repetitive textures with corals, white cloths for textureless regions.


\subsection{The optimal TS generation} \label{subsec:Qualitative}

To validate the proposed structure-aware information metric and the BO strategy, Fig. \ref{f6} (left) evaluates three distinct scenarios. The discrete sampling curves confirm that the optimal $\tau$ varies across different velocities and textures. The adjacent BO curves demonstrate that the surrogate model accurately predicts these optima. Qualitatively, the highlighted optimal TS frames exhibit the sharpest edge structures with minimal motion blur compared to other fixed $\tau$ values. 

Fig. \ref{f6} (right) illustrates the overall performance of the dual-stage optimization on the highly dynamic UWE S10 sequence. The time-series plot reveals an inverse correlation between the adapted $\tau$ (blue) and camera velocity (orange). As velocity drops in Stage \textcircled{2}, $\tau$ adaptively increases to accumulate sufficient events. During the sharp acceleration and subsequent deceleration in Stage \textcircled{3}, $\tau$ rapidly decreases to prevent blurring before rebounding. The 3D trajectory and its corresponding sampled TS frames further verify that our strategy consistently generates high-quality event representations throughout aggressive maneuvers.

\subsection{Adaptive Disparity-Guided Matching}

\begin{figure}[htbp]
    \centering
    \includegraphics[width=0.9\linewidth]{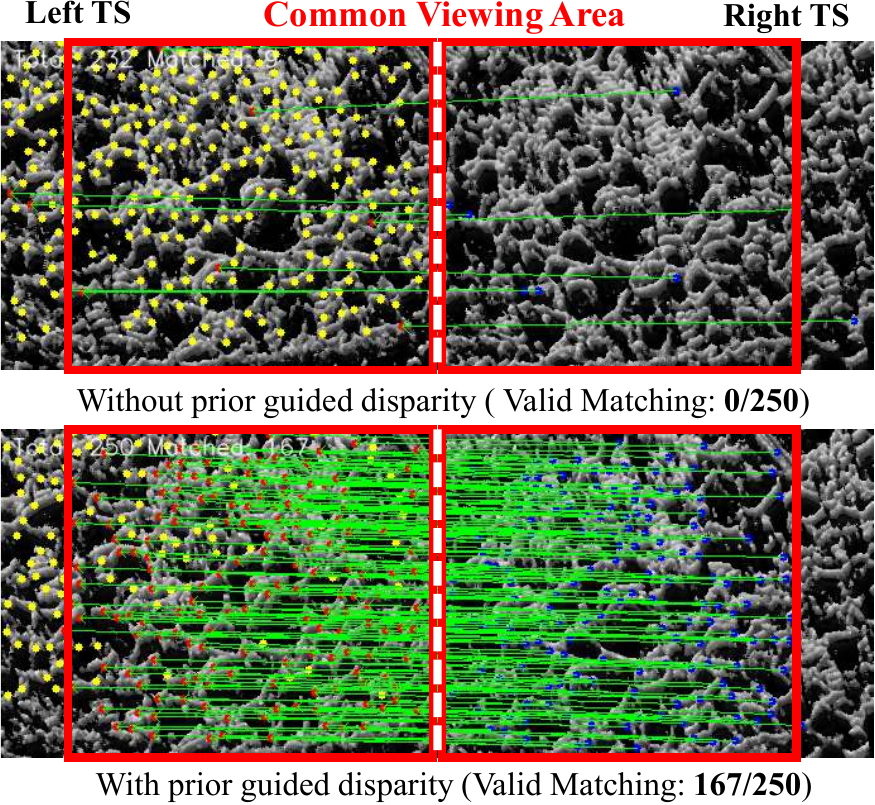}
    \captionsetup{justification=justified,singlelinecheck=false}
    \caption{Matching comparison under repetitive textures. Yellow points denote extracted left features, while red and blue points represent matched points in the left and right TS. Red boxes bound the common viewing area for valid candidates.}
    \label{f7}
\end{figure}

Fig. \ref{f7} evaluates the proposed adaptive disparity-guided stereo matching against the baseline LK optical flow in challenging, repetitive coral textures. The baseline method severely degenerates, establishing merely 9 mismatching pairs out of 250 feature points (top row). In stark contrast, by introducing a spatial offset prior to constrain the search range, our algorithm overcomes the spatial ambiguity, accurately establishing 167 reliable correspondences in common view area (bottom row). This improvement confirms the effectiveness of our data association method under large disparities and repetitive textures.

\begin{table*}[htpb]
\centering
\caption{Quantitative evaluation of ablation studies on ATE [m] using UWE dataset}
\label{t1}
\renewcommand{\arraystretch}{1.3}

\resizebox{\textwidth}{!}{%
\begin{tabular}{ccccccccccccccc}
\toprule
\multirow{3}{*}{\begin{tabular}[c]{@{}c@{}}Sequences \\ (characteristics)\end{tabular}} & \multicolumn{6}{c}{Original ESIO} & \multicolumn{6}{c}{ESIO with improved stereo tracking and triangulation} & \multicolumn{2}{c}{Ours} \\
\cmidrule(lr){2-7} \cmidrule(lr){8-13} \cmidrule(lr){14-15}
& \multicolumn{2}{c}{20ms} & \multicolumn{2}{c}{50ms} & \multicolumn{2}{c}{80ms} & \multicolumn{2}{c}{20ms} & \multicolumn{2}{c}{50ms} & \multicolumn{2}{c}{80ms} & \multicolumn{2}{c}{auto} \\
\cmidrule(lr){2-3} \cmidrule(lr){4-5} \cmidrule(lr){6-7} \cmidrule(lr){8-9} \cmidrule(lr){10-11} \cmidrule(lr){12-13} \cmidrule(lr){14-15}
& RMSE & S.D. & RMSE & S.D. & RMSE & S.D. & RMSE & S.D. & RMSE & S.D. & RMSE & S.D. & RMSE & S.D. \\
\midrule
S01(\textit{E.E.E}) & \underline{0.140} & \underline{0.045} & 0.174 & 0.057 & 0.188 & 0.063 & \textbf{0.110} & \textbf{0.035} & 0.138 & 0.043 & 0.149 & 0.046 & \underline{\textbf{0.102}} & \underline{\textbf{0.033}} \\ 
S02(\textit{E.E.E}) & \underline{0.134} & \underline{0.060} & 0.155 & 0.069 & 0.163 & 0.070 & \textbf{0.098} & \textbf{0.041} & 0.120 & 0.053 & 0.128 & 0.054 & \underline{\textbf{0.093}} & \underline{\textbf{0.037}} \\ 
S03(\textit{M.E.E}) & \underline{0.141} & 0.082 & 0.147 & \underline{0.072} & 0.185 & 0.089 & \textbf{0.085} & \textbf{0.036} & 0.100 & 0.043 & 0.125 & 0.054 & \underline{\textbf{0.082}} & \underline{\textbf{0.033}} \\ 
S04(\textit{H.E.E}) & 0.204 & 0.095 & \underline{0.195} & \underline{0.084} & 0.234 & 0.105 & \textbf{0.133} & \textbf{0.059} & 0.151 & 0.067 & 0.161 & 0.069 & \underline{\textbf{0.069}} & \underline{\textbf{0.031}} \\ 
S05(\textit{M.E.E}) & \underline{0.163} & \underline{0.063} & 0.191 & 0.070 & 0.218 & 0.095 & \textbf{0.114} & \textbf{0.045} & 0.129 & 0.052 & 0.149 & 0.062 & \underline{\textbf{0.109}} & \underline{\textbf{0.040}} \\ 
S06(\textit{M.E.E}) & 0.140 & 0.065 & \underline{0.114} & \underline{0.062} & 0.131 & 0.081 & \textbf{0.088} & \textbf{0.042} & 0.100 & 0.050 & 0.121 & 0.063 & \underline{\textbf{0.081}} & \underline{\textbf{0.039}} \\ 
S07(\textit{M.M.E}) & \underline{0.097} & \underline{0.054} & 0.098 & 0.056 & 0.117 & 0.070 & \textbf{0.084} & \textbf{0.044} & 0.091 & 0.045 & 0.099 & 0.056 & \underline{\textbf{0.075}} & \underline{\textbf{0.035}} \\ 
S08(\textit{$H^{+}$.H.E}) & \textit{failed} & \textit{failed} & 0.096 & 0.040 & \underline{0.089} & \underline{0.039} & \textbf{0.071} & \textbf{0.035} & \textbf{0.071} & 0.036 & 0.074 & \textbf{0.035} & \underline{\textbf{0.068}} & \underline{\textbf{0.033}} \\ 
S09(\textit{M.M.M}) & 0.229 & 0.150 & \underline{0.157} & \underline{0.087} & 0.233 & 0.137 & \textbf{0.124} & 0.060 & 0.129 & \textbf{0.058} & 0.145 & 0.060 & \underline{\textbf{0.121}} & \underline{\textbf{0.062}} \\ 
S10(\textit{H.H.H}) & 0.385 & 0.210 & \underline{0.239} & \underline{0.107} & \textit{failed} & \textit{failed} & \textbf{0.155} & \textbf{0.058} & 0.422 & 0.225 & 1.707 & 0.939 & \underline{\textbf{0.124}} & \underline{\textbf{0.044}} \\ 
\bottomrule
\multicolumn{15}{p{\textwidth}}{\footnotesize *The abbreviations in parentheses (e.g., \textit{E.E.E}) correspond to the difficulty levels of three metrics: robot motion, trajectory complexity, and visual texture complexity. Each metric is categorized into three levels: Easy (E), Medium (M), and High (H). \textit{$H^{+}$} means extremely hard. \underline{Underline} and \textbf{bold} denote the best fixed-$\tau$ results for Original ESIO and our method without adaptive $\tau$, respectively. \underline{\textbf{Bold underline}} highlights the overall best performance.}
\end{tabular}%
} 
\end{table*}

\subsection{Quantitative experiments} \label{subsec:Quantative}

Quantitative evaluations are conducted on the AQV and UWE datasets. We compare our method against the state-of-the-art stereo event-based system, ESVO2, and our baseline, ESVIO. Since our focus is on pure-event underwater applications, comparisons are strictly made against the pure-event mode (ESIO) of ESVIO. For a fair comparison, the valid depth ranges in ESIO are aligned with our configurations and the motion compensation function was disabled in our system.

\subsubsection{\textbf{Ablation experiment}} 

To quantitatively assess the individual contributions of the proposed modules, an ablation study is performed on the UWE dataset. Table \ref{t1} compares the Absolute Trajectory Error (ATE) of the original ESIO, the ESIO augmented with our robust stereo tracking and triangulation, and our full system with the best results being bolded. The baseline tests are evaluated at fixed $\tau$ values of $20$ms, $50$ms, and $80$ms. Results indicate that the robust triangulation module consistently outperforms the original system under identical $\tau$ settings. Furthermore, our full system achieves the best overall performance across all sequences. One specific fixed $\tau$ can approach the optimal result, but an excessively large $\tau$ significantly degrades localization accuracy, directly proving the necessity of dynamically selecting the optimal $\tau$. Notably, while the original ESIO suffers from tracking failures in certain sequences (S08), our proposed system operates stably without any failures, exhibiting superior robustness and stability in complex underwater environments.

\subsubsection{\textbf{Localization accuracy}} 

Table \ref{t2} presents the quantitative comparison among ESVO2, ESIO, and our proposed method on the UWE dataset. To ensure a competitive baseline, we report the results for ESIO using $\tau = 20$ms, which yields its highest overall accuracy. The results unequivocally demonstrate that our algorithm achieves the best performance across all sequences, yielding the lowest RMSE and Standard Deviation (S.D.) of the ATE. This indicates not only superior overall localization accuracy but also a highly concentrated error distribution, leading to strictly consistent trajectory estimation. 

Specifically, in S01--S08, our RMSE remains stable between $0.068$m and $0.11$m. In the more challenging S09 and S10, our method maintains an RMSE of approximately $0.12$m, delivering a nearly $50\%$ improvement over ESIO. This substantial gain stems from dynamically preserving the optimal TS structure and highly stable initialization. Conversely, ESVO2 completely fails in these underwater scenarios. This failure is primarily due to its reliance on excessive hyperparameter tuning, making it difficult to adapt to our aquatic environments.


\begin{table}[htbp]
\centering
\caption{Quantitative comparison of SOTA Event-Base methods on ATE [m] using UWE dataset}
\label{t2}
\setlength{\tabcolsep}{7pt}
\renewcommand{\arraystretch}{1.3}
\begin{tabular}{lcccccc}
\toprule
\multirow{2}{*}{Sequence} & \multicolumn{2}{c}{ESVO2} & \multicolumn{2}{c}{ESIO} & \multicolumn{2}{c}{Ours} \\
\cmidrule(lr){2-3} \cmidrule(lr){4-5} \cmidrule(lr){6-7}
& RMSE & S.D. & RMSE & S.D. & RMSE & S.D. \\
\midrule
S01 & \multicolumn{2}{c}{\textit{failed}} & 0.140 & 0.045 & \textbf{0.102} & \textbf{0.033} \\
S02 & \multicolumn{2}{c}{\textit{failed}} & 0.134 & 0.060 & \textbf{0.093} & \textbf{0.037} \\
S03 & \multicolumn{2}{c}{\textit{failed}} & 0.141 & 0.082 & \textbf{0.082} & \textbf{0.033} \\
S04 & \multicolumn{2}{c}{\textit{failed}} & 0.204 & 0.095 & \textbf{0.069} & \textbf{0.031} \\
S05 & \multicolumn{2}{c}{\textit{failed}} & 0.163 & 0.063 & \textbf{0.109} & \textbf{0.040} \\
S06 & \multicolumn{2}{c}{\textit{failed}} & 0.140 & 0.065 & \textbf{0.081} & \textbf{0.039} \\
S07 & \multicolumn{2}{c}{\textit{failed}} & 0.097 & 0.054 & \textbf{0.075} & \textbf{0.035} \\
S08 & \multicolumn{2}{c}{\textit{failed}} & \multicolumn{2}{c}{\textit{failed}} & \textbf{0.068} & \textbf{0.033} \\
S09 & \multicolumn{2}{c}{\textit{failed}} & 0.229 & 0.150 & \textbf{0.121} & \textbf{0.062} \\
S10 & \multicolumn{2}{c}{\textit{failed}} & 0.385 & 0.210 & \textbf{0.124} & \textbf{0.044} \\
\bottomrule

\end{tabular}
\end{table}

\subsubsection{\textbf{Generalization ability}} 

To comprehensively evaluate the generalization capability of the proposed system, additional experiments were conducted on the AQV dataset. As detailed in Table \ref{t3}, our method consistently secures the best performance even under severely degraded conditions, such as poor event data quality, extremely slow sensor motion, and highly texture-less scenes. These compelling results further validate the outstanding adaptability and robust generalization capacity of the proposed algorithm across diverse and challenging aquatic environments.


\begin{table}[htbp]
\centering
\caption{Generalization ability of Our Method on \textbf{AQ}uatic\textbf{V}ision dataset~\cite{peng2025aquaticvision}  [m]}
\label{t3}
\setlength{\tabcolsep}{5pt}
\renewcommand{\arraystretch}{1.3}
\begin{tabular}{lcccccc}
\toprule
\multirow{2}{*}{Sequences} & \multicolumn{2}{c}{ESVO2} & \multicolumn{2}{c}{ESIO} & \multicolumn{2}{c}{Ours} \\
\cmidrule(lr){2-3} \cmidrule(lr){4-5} \cmidrule(lr){6-7}
& RMSE & S.D. & RMSE & S.D. & RMSE & S.D. \\
\midrule
Scan\_with\_board* & \multicolumn{2}{c}{\textit{failed}} & 0.180 & 0.109 & \textbf{0.079} & \textbf{0.031} \\ 
Cross1\_with\_board & \multicolumn{2}{c}{\textit{failed}} & 0.066 & 0.034 & \textbf{0.057} & \textbf{0.023} \\%
Cross2\_no\_board & \multicolumn{2}{c}{\textit{failed}} & 0.065 & 0.031 & \textbf{0.056} & \textbf{0.026} \\%
Loop1\_with\_board* & \multicolumn{2}{c}{\textit{failed}} & 0.087 & 0.052 & \textbf{0.058} & \textbf{0.027} \\ 
Loop2\_no\_board & \multicolumn{2}{c}{\textit{failed}} & 0.096 & 0.030 & \textbf{0.082} & \textbf{0.028} \\ %
Dark1\_with\_board & \multicolumn{2}{c}{\textit{failed}} & 0.246 & 0.065 & \textbf{0.235} & \textbf{0.063} \\%
Dark2\_with\_board* & \multicolumn{2}{c}{\textit{failed}} & 0.069 &  0.032 & \textbf{0.056} & \textbf{0.026} \\%
HDR & \multicolumn{2}{c}{\textit{failed}} & 0.178 & 0.073 & \textbf{0.131} & \textbf{0.041} \\
\bottomrule
\multicolumn{7}{l}{\footnotesize $^*$ denotes sequences easy to cause system failure.}
\end{tabular}
\end{table}

\subsection{Real-time performance analysis} 

To evaluate computational efficiency, all experiments are conducted on a computer equipped with an Intel Core i7-10875H processor, strictly relying on CPU computation without any GPU acceleration. Table \ref{t4} summarizes the average execution times of the core proposed modules across multiple test sequences.

Although the BO prediction requires $46.415$\,ms, it is executed exclusively during the pre-SfM cold start phase. Therefore, it introduces absolutely zero latency to the continuous online tracking. During normal operation, the asynchronous online search takes approximately $15.844$\,ms. Because this process is entirely decoupled and isolated in an independent background thread, it never blocks the main tracking pipeline. Furthermore, the lightweight design of our algorithms ensures that a single evaluation of the information metric and the adaptive disparity calculation consume only $0.740$\,ms and $0.020$\,ms, respectively---both well under $1$\,ms. These highly optimized processing times demonstrate that the proposed pure-event stereo SLAM system achieves exceptional real-time performance, making it highly suitable for deployment on resource-constrained onboard platforms.

\begin{table}[htbp]
\centering
\caption{Real-time performance of our algorithm [ms]}
\label{t4}
\setlength{\tabcolsep}{9.5pt}
\renewcommand{\arraystretch}{1.3}
\begin{tabular}{ccccccc}
\toprule
 & \makecell{Information \\ Metric} & \makecell{BO \\ Prediction} & \makecell{Online \\ Search} & \makecell{ Adaptive \\ disparity}  \\
\midrule
Time & 0.740 & 46.415 & 15.844 & 0.020  \\
\bottomrule
\end{tabular}
\end{table}

\section{Conclusion}

This paper presents a highly accurate and robust pure-event stereo SLAM system tailored for underwater environments. By designing a structure-aware information metric for Time Surfaces alongside a dual-stage adaptive optimization strategy, the system guarantees real-time optimal TS generation, effectively mitigating representation degradation under varying motions and textures. Additionally, a median disparity prior and a "latest-observation-first" depth propagation mechanism resolve data association failures under large disparities and repetitive scenes. Extensive evaluations on public benchmarks and our newly released UWE dataset confirm that the proposed algorithm outperforms state-of-the-art methods in accuracy and stability while preserving real-time efficiency. Future work will focus on event-based loop closure and deployment on autonomous underwater vehicles.

\bibliographystyle{ieeetr}
\bibliography{paper}

\end{document}